\documentclass[10pt,letterpaper]{article}

\usepackage{cogsci}
\cogscifinalcopy

\usepackage[style=apa,natbib=true,annotation=false,]{biblatex}
\usepackage{float}
\usepackage{graphicx}
\usepackage{booktabs}
\usepackage{hyperref}
\usepackage{cleveref}
\usepackage{fontawesome5}

\title{Revisiting Real-Time Digging-In Effects: No Evidence from NP/Z Garden-Paths}

\author{\mbox{Amani Maina-Kilaas (amanirmk@mit.edu)}}
\author{\mbox{Roger Levy}}
\affil{Department of Brain and Cognitive Sciences, MIT}

\begin{document}

\maketitle

\begin{abstract}
Digging-in effects, where disambiguation difficulty increases with longer ambiguous regions, have been cited as evidence for self-organized sentence processing, in which structural commitments strengthen over time. In contrast, surprisal theory predicts no such effect unless lengthening genuinely shifts statistical expectations, and neural language models appear to show the opposite pattern. Whether digging-in is a robust real-time phenomenon in human sentence processing---or an artifact of wrap-up processes or methodological confounds---remains unclear. We report two experiments on English NP/Z garden-path sentences using Maze and self-paced reading, comparing human behavior with predictions from an ensemble of large language models. We find no evidence for real-time digging-in effects. Critically, items with sentence-final versus nonfinal disambiguation show qualitatively different patterns: positive digging-in trends appear only sentence-finally, where wrap-up effects confound interpretation. Nonfinal items---the cleaner test of real-time processing---show reverse trends consistent with neural model predictions.

\textbf{Keywords:} digging-in effects; sentence processing; garden-path sentences; surprisal theory; large language models
\end{abstract}

\section{Introduction}

Readers sometimes commit to a syntactic analysis that later proves incorrect. In a sentence like \textit{``Before the recruiters could adapt the policy changed''}, the noun phrase \textit{the policy} is initially interpreted as the object of \textit{adapt}---but the verb \textit{changed} forces reanalysis, revealing that \textit{the policy} actually begins a new clause. The resulting difficulty, evident in longer reading times at the disambiguating word, is the garden-path effect~\citep{bever1970structures, frazier1982correcting}.

A related phenomenon focuses on what happens when the ambiguous region is lengthened. In \textit{``Before the recruiters could adapt the policy for hiring changed''}, the modifier \textit{for hiring} extends the period during which readers maintain the incorrect analysis. If lengthening increases disambiguation difficulty---a digging-in effect---this suggests that structural commitments strengthen over time~\citep{Tabor2004EvidenceFS,frazier1982correcting,warner1987context}.

Dynamical, self-organizing models of parsing~\citep{Tabor2004EvidenceFS,TABOR2004355,TABOR1999491} naturally predict digging-in: syntactic analyses have continuous-valued activation that increases through self-reinforcement during processing. The longer a parse is maintained, the harder it becomes to revise. Surprisal theory~\citep{Hale2001APE,Levy2008ExpectationbasedSC}, by contrast, ties processing difficulty to the information content of incoming words. Under this account, lengthening alone should not increase difficulty unless it changes the probability distribution over upcoming continuations. Empirical confirmation that digging-in effects are a robust \textit{real-time} phenomenon---directly reflecting commitment during parsing---that does not reduce to changes in surprisal is of theoretical importance for constraining the space of viable processing models. Digging-in effects that do not result from statistical expectations would challenge purely expectation-based accounts, while the lack thereof would challenge models that predict time-dependent strengthening of commitments. 

In testing surprisal theory, we need to estimate the comprehender's subjective probability of a word in context, which we assume is calibrated to the language statistics of their environment. Neural language models are currently our best scalable estimate of this probability: they are fully incremental and totally parallel, and their surprisal robustly predicts reading times~\citep{cory2024large, GotliebWilcox2023TestingTP} and neural responses~\citep{Michaelov2023StrongPL}. Crucially, these models continue to provide accurate estimates even in the low range of probabilities, which is where the words involved in garden-path disambiguation fall.

The current empirical work with neural language models complicates the picture on digging-in effects. \citet{Futrell2019NeuralLM} found that recurrent neural networks---which implicitly maintain all parses in parallel---exhibit small reverse digging-in effects, where lengthening slightly reduces disambiguation difficulty. If human processing difficulty tracks surprisal, we may actually expect slight reverse effects in behavioral data---creating diverging predictions for the fully-parallel surprisal theory and self-reinforcing dynamical models.

We present two experiments testing whether digging-in is a robust real-time phenomenon in English NP/Z garden-path sentences. We employ both the Maze task~\citep{Forster2009TheMT,Boyce2020MazeME}, which provides highly localized measurements of word-by-word processing difficulty, and self-paced reading~\citep{Just1980}, which replicates the methodology of the original study. For each, we compare human behavior with response times predicted by an ensemble of large language models, operationalizing surprisal theory.

\begin{table*}[htbp]
    \centering
    \begin{tabular}{ll}
    \toprule
    Ambiguity & Stimuli \\
    \midrule
        ambiguous & Before the recruiters could adapt the policy (for hiring) \textbf{changed} (for a second time).\\
        unamb (object) & Before the recruiters could adapt their strategies the policy (for hiring) \textbf{changed} (for a second time).\\
        unamb (comma) & Before the recruiters could adapt, the policy (for hiring) \textbf{changed} (for a second time).\\
    \bottomrule
    \end{tabular}
    \caption{Manipulations for a self-paced reading item with critical word ``changed''. \textbf{Ambiguity}: ambiguous vs. unambiguous. \textbf{Resolution:} resolves ambiguity through a new NP (object) or inserting a comma (comma); typically collapsed in analyses. \textbf{Length}: short vs.~long (adds ``for hiring''). \textbf{Finality}: final vs.~nonfinal (adds ``for a second time''). Maze items use the same levels but vary resolution and finality between-items, with resolution balanced but primarily nonfinal items (22 of 30).}
    \label{tab:examples}
\end{table*}

Critically, in self-paced reading, we examine a factor not previously explored: whether the disambiguating word occurs sentence-finally or sentence-medially. \citet{Tabor2004EvidenceFS} used exclusively sentence-final critical words to concentrate the disambiguation effect. However, sentence-final reading times are confounded with wrap-up processes~\citep{Meister2022AnalyzingWE} that may interact in unclear ways with garden-path recovery. Sentence-nonfinal items provide a cleaner test of real-time processing. Our results reveal no evidence for digging-in in nonfinal positions---and suggest that the original finding may reflect a poorly-understood interaction with sentence wrap-up rather than increased structural commitment during parsing.

\section{General Methods}

Following \citet{Tabor2004EvidenceFS}, we test for digging-in effects in English NP/Z garden-paths, where a noun phrase following an optionally-transitive verb is temporarily ambiguous. In the ambiguous version of the sentence, it is unclear whether the noun phrase is the object of the verb until a late disambiguation from a second verb (henceforth \textbf{critical word}), causing a garden-path. In the unambiguous version, the garden-path is eliminated by including an early disambiguating region, where either a comma marks the clause boundary or the optionally-transitive verb takes a different object. We compare the garden-path between short and long sentences, where the long version includes a post-nominal modifier in the ambiguous noun phrase. Digging-in predicts a larger garden-path effect in long sentences, quantified by a positive interaction between ambiguity and length at the critical word.

All items were designed with contrasts for \textbf{ambiguity} and \textbf{length}. Items also varied in two additional ways: \textbf{resolution}, the type of disambiguation in unambiguous versions (a comma marking a clause boundary or the verb taking a different object), and \textbf{finality}, the location of the critical word (sentence-final or sentence-nonfinal). In Experiment 1, resolution and finality varied naturally across items. In Experiment 2, we formalized them as within-item contrasts. An example item is provided in \Cref{tab:examples} with all four factors.

\subsection{Human Experiments}

We conducted two human experiments, using Maze~\citep{Boyce2020MazeME,Forster2009TheMT} and self-paced reading~\citep{Just1980}, to address potential methodological limitations of the prior work. For both experiments, we recruited participants via Prolific, screening for English as first language and United States residence. Participants were compensated at approximately \$15/hr. Before analyses, participants were excluded based on accuracy in filler items and trials were excluded based on extreme values and sentence comprehension; additional details are provided under each experiment.

\subsection{Surprisal Theory Predictions}

For each experiment, we compare the empirical data with response times (RTs) predicted by an ensemble of 16 large language models (LLMs) from 5 model families: \textbf{GPT-2} (small, medium, large; \citealt{radford2019language}), \textbf{Pythia} (70m, 160m, 1b, standard and deduped versions; \citealt{biderman2023pythia}), \textbf{Gemma-3} (1b and 4b pretrained; \citealt{gemma2025}), \textbf{Qwen 2.5} (0.5b, 1.5b, 3b, 7b; \citealt{qwen2024}), and \textbf{Mistral} (7b-v0.3; \citealt{jiang2023mistral7b}).

For each LLM, we predict RTs by fitting ordinary least squares regression to mean word RTs in filler items (excluding the first and final word), using surprisal (obtained via the \texttt{minicons} interface to \texttt{transformers}; \citealt{misra2022minicons, wolf2020transformers}), log-transformed word frequency (via \texttt{wordfreq}; \citealt{wordfreq}), and word length as predictors, along with these same features for the two previous words to account for spillover. We then generate predictions for critical items. These predictions are then aggregated by treating individual LLMs as participants in subsequent analyses, so as to reflect surprisal theory above individual language model specifics.

\subsection{Statistical Analysis}

We analyzed both empirical and predicted response times using Bayesian mixed-effects models in \texttt{brms}~\citep{Burkner2017brms} with default uninformative priors; each was run with 5 chains for 5000 iterations. Models always used maximal random-effects structures~\citep{barr2013maximal} with by-item and by-participant (or by-LLM) random slopes and intercepts. We used sum coding for experimental factors, where levels are coded as $+$1 and $-$1 such that coefficient estimates represent deviations from the grand mean ($+$1: ambiguous, long, comma, final). This ensures effects are estimated using all data rather than only baseline conditions. For summary statistics, we report the posterior median ($m$) and probability of direction ($p_d$), computed using \texttt{bayestestR}~\citep{bayestest}. The raw effect size (difference between conditions) is \textit{twice} the posterior median due to the coding scheme. We consider effects ``statistically significant'' when \mbox{$p_d > 97.5\%$}, which corresponds conceptually to a two-tailed \mbox{$p < 0.05$}. Data and analysis scripts are available: \faGithub\;\,\href{https://github.com/amanirmk/digging-in}{\texttt{amanirmk/digging-in}}.

\section{Experiment 1: Maze}

\citet{Tabor2004EvidenceFS} used self-paced reading, which can show meaningful spillover---where processing difficulty extends to subsequent words~\citep{Smith2013TheEO}. To address this, we used the Maze task~\citep{Forster2009TheMT,Boyce2020MazeME}, which provides highly localized measurements of processing difficulty~\citep{Boyce2023AmazeON}. In Maze, participants are presented with two words at each position and must select the one that properly continues the sentence. The other word is a distractor, automatically generated using \citet{gulordava-etal-2018-colorless}, designed to be incompatible with the context. By forcing full processing of each word before proceeding, Maze reduces spillover effects and concentrates difficulty at the location where disambiguation occurs.

\subsection{Additional Details}

We designed 30 critical items manipulating ambiguity and length. Each participant completed 42 sentences: 2 of 5 practice items, 25 of 35 fillers, and 15 of 30 critical items in a Latin square design. We recruited 191 participants and excluded those who scored below 80\% accuracy when selecting words in filler items (excluding the first word in each sentence), leaving 167. We then excluded trials (a complete sentence) containing any RTs less than 100ms or above 5000ms (again excluding the first word) as these are unlikely to reflect genuine language processing. In RT analyses, we filtered for comprehension of the prior context by including only trials where participants made correct selections up to \textit{but not including} the critical word.

\subsection{Results}

\begin{figure}[htbp]
    \centering
    \includegraphics[width=\linewidth]{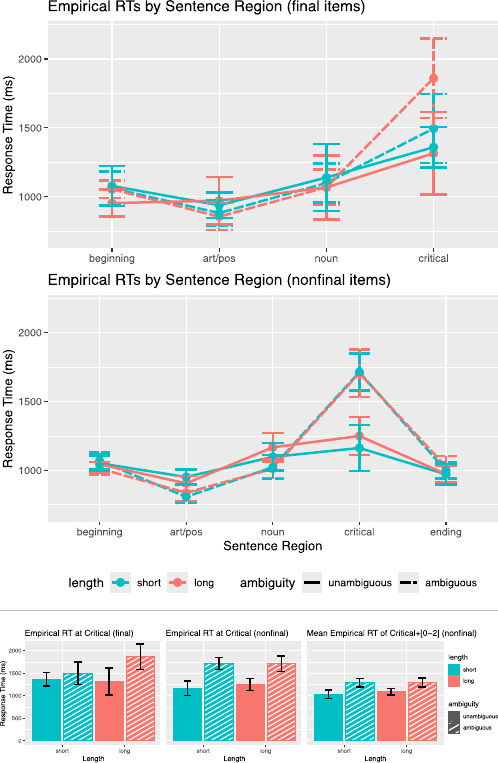}
    \caption{Empirical response times for Experiment 1 (Maze), split by sentence-finality. Top panels show mean word RT by sentence region (omitting regions not in all conditions). Bottom panels show the mean critical word RT, with the right-most averaging RTs within a plausible spillover region. Error bars reflect 95\% confidence intervals around by-item means.}
    \label{fig:maze_line}
\end{figure}

\begin{figure}[htbp]
    \centering
    \includegraphics[width=\linewidth]{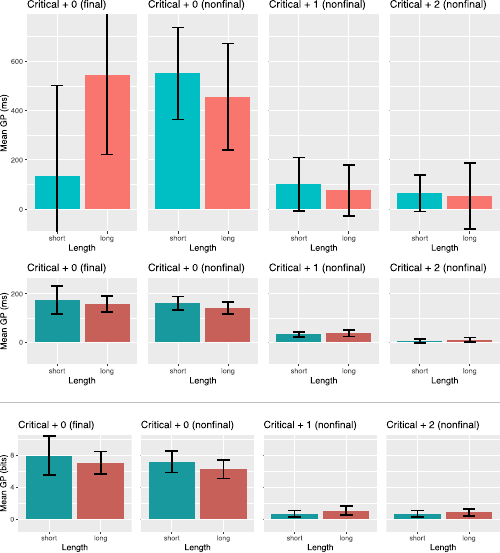}
    \caption{Mean garden-path effect in Experiment 1 (Maze), split by sentence-finality. Top row shows empirical data, middle shows predicted data; bottom shows surprisal for reference. Error bars reflect 95\% confidence intervals around by-item means, but readers should rely on the mixed-effects models for assessing significance due to better variance attribution.}
    \label{fig:maze_gp}
\end{figure}

\Cref{fig:maze_line} displays the empirical response times over the course of the sentence, while \Cref{fig:maze_gp} isolates the garden-path effect to compare short and long conditions for humans and LLMs.

\subsubsection{Empirical.}

Altogether, human participants showed a robust garden-path effect at the critical word (\mbox{$m = 225.79$ms}, \mbox{$p_d = 100$\%}), but no evidence of digging-in (\mbox{$m = 11.71$ms}, \mbox{$p_d = 70.83\%$}). However, we observed substantial differences based on resolution and finality. Comma-disambiguated items produced substantially smaller garden-path effects (\mbox{$m = 80.03$ms}, \mbox{$p_d = 99.61\%$}) than object-disambiguated items (\mbox{$m = 374.42$ms}, \mbox{$p_d = 100\%$}). More importantly for digging-in, sentence-final items showed a positive interaction approaching significance (\mbox{$m = 93.64$ms}, \mbox{$p_d = 97.38\%$}), while sentence-nonfinal items exhibited a reverse trend (\mbox{$m = -18.24$ms}, \mbox{$p_d = 76.30\%$}). Nonfinal items showed modest spillover to the following word (ambiguity effect: \mbox{$m = 44.44$ms}, \mbox{$p_d = 99.07\%$}), where the reverse digging-in trend persisted (\mbox{$m = -3.61$ms}, \mbox{$p_d = 58.81\%$}). There were no statistically significant effects in word selection accuracy.

\subsubsection{Predicted.}

LLM-predicted response times showed a robust garden-path effect (\mbox{$m = 77.32$ms}, \mbox{$p_d = 100\%$}) and a statistically significant reverse digging-in trend (\mbox{$m = -4.54$ms}, \mbox{$p_d = 99.25\%$}). In contrast to human behavior, predictions showed comparable garden-path effects for both comma-disambiguated items (\mbox{$m = 66.24$ms}, \mbox{$p_d = 100\%$}) and object-disambiguated items (\mbox{$m = 88.66$ms}, \mbox{$p_d = 100\%$}). Similarly, LLMs exhibited consistent reverse digging-in trends in both sentence-final (\mbox{$m = -3.88$ms}, \mbox{$p_d = 82.22\%$}) and sentence-nonfinal items (\mbox{$m = -4.84$ms}, \mbox{$p_d = 97.82\%$}), showing no modulation by finality.

\subsection{Discussion}

Contrary to \citet{Tabor2004EvidenceFS}, we found no statistically significant evidence for digging-in effects in humans. However, resolution and finality---factors that varied naturally among our items---proved impactful. Commas were substantially less effective at resolving ambiguity than objects, suggesting noisy-channel processing or task-specific considerations. More critically, sentence-final and sentence-nonfinal items showed qualitatively different patterns. Since sentence-final response times are confounded with wrap-up effects and are often excluded from analyses~\citep{Meister2022AnalyzingWE}, nonfinal items provide the cleaner test of \textit{real-time} digging-in---which showed the reverse trends predicted by LLMs. Our LLM results replicate both the reverse digging-in reported by \citet{Futrell2019NeuralLM} and the systematic underprediction of garden-path effects documented by \citet{Huang2024LargescaleBY}.

\section{Experiment 2: Self-Paced Reading}

Experiment 1 revealed that resolution type and finality substantially affected behavioral patterns. Since \citet{Tabor2004EvidenceFS} used only object-disambiguated sentence-final items, these factors may be critical to understanding their findings. To test this systematically, we returned to the self-paced reading (SPR) paradigm~\citep{Just1980} used in the original study. In non-cumulative self-paced reading, participants press a button to reveal each word of a sentence one at a time, with previous and future words masked.

\subsection{Additional Details}

We designed 36 critical items with ambiguity, length, resolution, and finality all manipulated within-items. Each participant completed 48 sentences: 3 of 5 practice items, 27 of 35 fillers, and 18 of 36 critical items in a Latin square design. Unlike the forced-choice Maze task, self-paced reading allows participants to advance quickly without reading carefully. To address this, we included binary-choice comprehension questions after each sentence and offered a 15\% payment bonus for achieving at least 80\% accuracy on non-practice questions. We recruited 233 participants and excluded those who scored below 80\% accuracy on filler items, retaining 225. As with Experiment 1, we removed trials containing any RTs below 100ms or above 5000ms (excluding sentence-initial words). In RT analyses, we filtered for comprehension by including only trials where participants answered the comprehension question correctly.

\subsection{Results}

\begin{figure}[htbp]
    \centering
    \includegraphics[width=\linewidth]{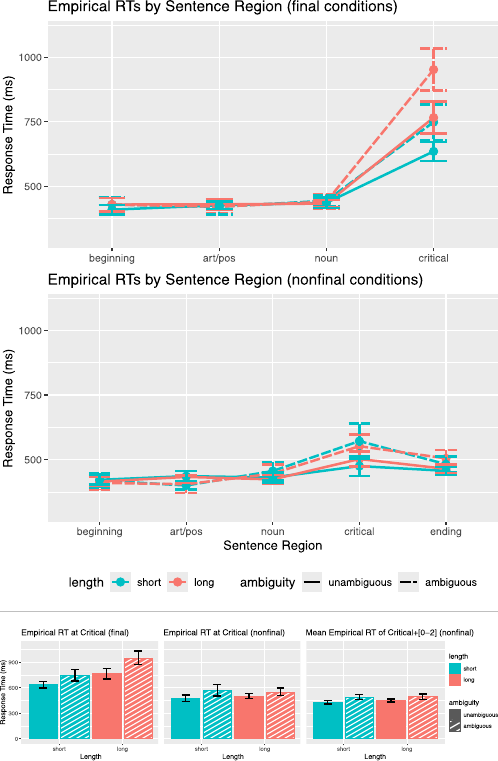}
    \caption{Empirical response times for Experiment 2 (SPR), split by sentence-finality. Top panels show mean word RT by sentence region (omitting regions not in all conditions). Bottom panels show the mean critical word RT, with the right-most averaging RTs within a plausible spillover region. Error bars reflect 95\% confidence intervals around by-item means.}
    \label{fig:spr_line}
\end{figure}

\begin{figure}[htbp]
    \centering
    \includegraphics[width=\linewidth]{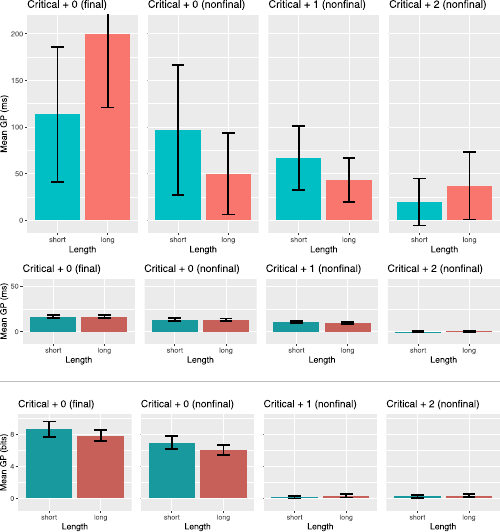}
    \caption{Mean garden-path effect in Experiment 2 (SPR), split by sentence-finality. Top row shows empirical data, middle shows predicted data; bottom shows surprisal for reference. Error bars reflect 95\% confidence intervals around by-item means, but readers should rely on the mixed-effects models for assessing significance due to better variance attribution.}
    \label{fig:spr_gp}
\end{figure}

\Cref{fig:spr_line} displays the empirical response times over the course of the sentence and \Cref{fig:spr_gp} isolates the garden-path effect to compare short and long conditions for humans and LLMs.

\subsubsection{Empirical.}

We found no substantial difference between comma and object disambiguation (\mbox{$m = 8.58$ms}, \mbox{$p_d = 83.93\%$}) and therefore collapsed the two unambiguous conditions for subsequent analyses. In sentence-nonfinal conditions, participants showed a robust garden-path effect at the critical word (\mbox{$m = 34.45$ms}, \mbox{$p_d = 99.77\%$}) but a reverse digging-in trend (\mbox{$m = -8.68$ms}, \mbox{$p_d = 82.91\%$}). This pattern continued at spillover, where the ambiguity effect remained significant (\mbox{$m = 26.77$ms}, \mbox{$p_d = 100\%$}) alongside a persistent reverse digging-in trend (\mbox{$m = -6.04$ms}, \mbox{$p_d = 89.06\%$}). In sentence-final conditions, the garden-path effect was substantially larger (\mbox{$m = 75.87$ms}, \mbox{$p_d = 100\%$}) with a positive digging-in trend (\mbox{$m = 19.69$ms}, \mbox{$p_d = 92.62\%$}). Analyzing all conditions together revealed robust two-way interactions between ambiguity and finality (\mbox{$m = 19.65$ms}, \mbox{$p_d = 98.43\%$}) and between length and finality (\mbox{$m = 36.43$ms}, \mbox{$p_d = 100\%$}), but no ambiguity by length interaction (\mbox{$m = 5.66$ms}, \mbox{$p_d = 76.63\%$}). The three-way interaction did not reach significance (\mbox{$m = 14.42$ms}, \mbox{$p_d = 96.63\%$}). There were no statistically significant effects in comprehension question accuracy.

\subsubsection{Predicted.}

LLM-predicted response times showed robust garden-path effects in both sentence-nonfinal (\mbox{$m = 6.46$ms}, \mbox{$p_d = 100\%$}) and sentence-final conditions (\mbox{$m = 8.24$ms}, \mbox{$p_d = 100\%$}), but no evidence of digging-in in either (nonfinal: \mbox{$m = -0.05$ms}, \mbox{$p_d = 60.14\%$}; final: \mbox{$m = -0.02$ms}, \mbox{$p_d = 53.66\%$}). At spillover in nonfinal items, predictions showed a continued ambiguity effect (\mbox{$m = 4.91$ms}, \mbox{$p_d = 100\%$}) alongside a small but reliable reverse digging-in trend (\mbox{$m = -0.31$ms}, \mbox{$p_d = 98.60\%$}). Raw LLM surprisal showed a clearer pattern, with reliable reverse digging-in at the critical word in both nonfinal (\mbox{$m = -0.22$ bits}, \mbox{$p_d = 99.35\%$}) and final conditions (\mbox{$m = -0.21$ bits}, \mbox{$p_d = 98.86\%$}).

\subsection{Discussion}

Experiment 2 recovers the qualitative discrepancy observed in Experiment 1: humans show digging-in trends in sentence-final conditions but reverse trends in nonfinal conditions. Critically, we find converging evidence that sentence-finality moderates the pattern, though the three-way interaction remains marginal (\mbox{$p_d = 96.63\%$}). While the processing of sentence-final words remains poorly understood, these response times may partially reflect post-sentence difficulty, consistent with the observation that digging-in effects are most robustly attested in metalinguistic ratings~\citep{Tabor2004EvidenceFS,Levy2008ModelingTE}. In nonfinal conditions, human behavior aligns qualitatively with language statistics, which predict that longer garden-paths should be easier to resolve. The reason for this pattern remains unclear.

Notably, LLM predictions were substantially less conclusive in self-paced reading compared to Maze, highlighting a methodological advantage of Maze for comparative studies between humans and language models. In Maze, surprisal dominates the relationship with response time relative to log-frequency and word length, while in self-paced reading the effect of surprisal is weaker and further complicated by spillover.\looseness=-1

The effect of resolution type observed in Experiment 1 did not replicate here, suggesting that the finding was primarily task-specific. This is unsurprising given that the Maze task arguably de-emphasizes the comma disambiguation: commas appear in both the distractor and correct continuation, rendering punctuation irrelevant to participants' choices in the moment. Future work using Maze should avoid relying heavily on punctuation when testing psycholinguistic phenomena.

\section{Adequacy of LLM-Predicted Response Times}

Surprisal theory, operationalized through LLMs, has proven remarkably successful at predicting human reading times across a range of constructions \citep{cory2024large,GotliebWilcox2023TestingTP}. However, recent work  demonstrates systematic underprediction when structural expectations are violated \citep{Huang2024LargescaleBY,vanSchijndel2020SingleStagePM,wilcox-etal-2021-targeted}. Our data are well positioned to contribute to this conversation. For previous analyses, we fit a linear relationship between response times and LLM surprisal using filler items, then used this to generate predictions for critical items. We now ask: how well does this relationship generalize, and is it sensitive to experimental conditions?

\begin{figure}[htbp]
    \centering
    \includegraphics[width=\linewidth]{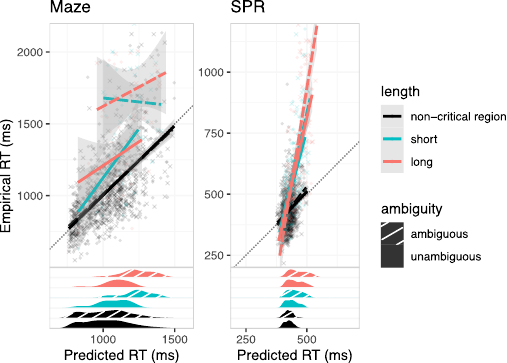}
    \caption{Empirical vs.~LLM-predicted response times in critical items. LLMs underpredict difficulty in disambiguating regions while accurately estimating in other sentence regions.}
    \label{fig:scatter}
\end{figure}

\Cref{fig:scatter} plots empirical vs.~predicted RTs. Each point represents the mean RT for a word (or set of words for SPR; see below) in a given item and condition. We split the data into non-critical words and critical words in short vs.~long conditions, then by ambiguity. A linear best-fit is shown for each subset.\looseness=-1

\subsection{Maze}

LLMs accurately predict response times for non-critical regions (mean residual \mbox{$\bar{r} = 4.05$ms}, \mbox{RMSE $= 199.61$ms}), with the best-fit line falling almost perfectly on \mbox{$y=x$}. For critical words in unambiguous conditions, predictions are accurate when the ambiguity is resolved by an object (\mbox{$\bar{r} = -3.05$ms}, \mbox{RMSE $= 161.62$ms}) but not when resolved by a comma (\mbox{$\bar{r} = 331.15$ms}, \mbox{RMSE $= 462.78$ms}), highlighting the limitation of comma-based disambiguation in Maze.

Critical words in ambiguous conditions are substantially underpredicted (\mbox{$\bar{r} = 468.95$ms}, \mbox{RMSE $= 587.22$ms}). Relevant to the discussion of digging-in, restricting to nonfinal conditions, critical words in long conditions (\mbox{$\bar{r} = 513.15$ms}, \mbox{RMSE $= 648.63$ms}) are predicted no worse than in short conditions (\mbox{$\bar{r} = 522.40$ms}, \mbox{RMSE $= 601.60$ms}).

\subsection{Self-Paced Reading}

We apply additional preprocessing before analyzing self-paced reading results to account for spillover effects. Because processing difficulty often manifests across multiple words following a critical region, we expand our definition of the critical region to include plausible spillover positions. Specifically, we average response times within 3-word chunks (i.e., \mbox{Critical$-$[3--1]}, \mbox{Critical$+$[0--2]}) for both participants and models before aggregating, and discard words beyond \mbox{Critical$+$2}.

LLMs accurately predict response times for non-critical regions (\mbox{$\bar{r} = 7.23$ms}, \mbox{RMSE $= 73.79$ms}). For critical words in unambiguous conditions, predictions are fairly accurate in nonfinal conditions (\mbox{$\bar{r} = 16.90$ms}, \mbox{RMSE $= 70.49$ms}) but not final conditions (\mbox{$\bar{r} = 232.30$ms}, \mbox{RMSE $= 300.78$ms}), suggesting wrap-up effects not captured by surprisal. Within nonfinal conditions, predictions are better when ambiguity is resolved by an object (\mbox{$\bar{r} = 7.10$ms}, \mbox{RMSE $= 57.65$ms}) than by a comma (\mbox{$\bar{r} = 26.72$ms}, \mbox{RMSE $= 81.33$ms}), perhaps a subtle indication of noisy-channel processing.

Critical words in ambiguous conditions are underpredicted, with a much larger mismatch in final (\mbox{$\bar{r} = 362.42$ms}, \mbox{RMSE $= 428.10$ms}) than nonfinal positions (\mbox{$\bar{r} = 61.06$ms}, \mbox{RMSE $= 105.82$ms}). Turning to digging-in and restricting to nonfinal conditions, long conditions (\mbox{$\bar{r} = 59.04$ms}, \mbox{RMSE $= 102.53$ms}) are again predicted no worse than short conditions (\mbox{$\bar{r} = 63.07$ms}, \mbox{RMSE $= 109.00$ms}).

\section{Conclusion}

Across two experiments, we found no evidence that digging-in effects are a robust real-time phenomenon in English NP/Z garden-path sentences. While humans showed robust garden-path effects in both Maze and self-paced reading, the ambiguity-by-length interaction was absent.

Positive digging-in trends emerged only in sentence-final conditions, where wrap-up processes confound interpretation. In nonfinal conditions, the trend reversed, aligning with surprisal theory predictions from recurrent neural networks \citep{Futrell2019NeuralLM} and LLMs. Our work indicates that ambiguity and length each separately interact with finality---not with each other---suggesting that the original finding of \citet{Tabor2004EvidenceFS}, which used exclusively sentence-final critical words, may not reflect increased structural commitment during parsing.

The analysis of LLM predictive adequacy further supports this picture: while critical words are underpredicted in ambiguous conditions (reinforcing \citet{Huang2024LargescaleBY}), this underprediction is no greater for long than for short conditions. Altogether, these findings challenge dynamical systems models that predict time-dependent strengthening and provide support for rational models that follow language statistics.

Several limitations should be mentioned. First, while we examined NP/Z garden-paths following the original study, this is only one grammatical construction. It is possible that strongly difficult garden-paths like NP/Z disrupt the normal processes required for digging-in effects. Second, our empirical data are statistically inconclusive---we are unable to clearly demonstrate that humans exhibit reverse digging-in effects as neural-based surprisal theory predicts. Third, sentence-final items did show consistent positive trends across both experiments, and understanding why wrap-up effects interact with ambiguity and length would itself be theoretically informative.

More broadly, our results highlight the importance of distinguishing real-time from post-sentence behavioral measures when testing theories of incremental processing. For real-time measures, sentence-final positions should be treated with due caution---or excluded entirely---since the goal in this case is to isolate \textit{incremental} sentence processing difficulty.

\section{Acknowledgments}

We thank the anonymous HSP 2026 reviewers for their helpful feedback. This work was supported by a grant from the Simons Foundation to the Simons Center for the Social Brain at MIT. Additionally, AMK is supported by the Fannie and John Hertz Foundation and an MIT Dean of Science Fellowship.

\printbibliography

\end{document}